\documentclass[10pt,twocolumn,letterpaper]{article}

\usepackage{cvpr}              % To produce the CAMERA-READY version
\usepackage[accsupp]{axessibility} 
\usepackage[dvipsnames]{xcolor}

\definecolor{cvprblue}{rgb}{0.21,0.49,0.74}
\usepackage[pagebackref,breaklinks,colorlinks,citecolor=cvprblue]{hyperref}
\usepackage{multirow}

\usepackage{array}
\newcolumntype{?}{!{\vrule width 1.5pt}}

\def\paperID{8190} % *** Enter the Paper ID here
\def\confName{CVPR}
\def\confYear{2024}

\title{Boosting Object Detection with Zero-Shot Day-Night Domain Adaptation}

\author{Zhipeng Du\thanks{Work partially done during internship at Huawei
London Research.\\ \indent\hspace{1mm}$^\dagger$Corresponding author}\\
\small Department of Informatics\\
\small King's College London\\
{\tt\small zhipeng.du@kcl.ac.uk}
\and
Miaojing Shi$^\dagger$\\
\small College of Electronic and Information Engineering\\
\small Tongji University\\
{\tt\small mshi@tongji.edu.cn}
\and
Jiankang Deng\\
\small Department of Computing\\
\small Imperial College London\\
{\tt\small j.deng16@imperial.ac.uk}
}

\begin{document}
\maketitle
\begin{abstract}
Detecting objects in low-light scenarios presents a persistent challenge, as detectors trained on well-lit data exhibit significant performance degradation on low-light data due to low visibility. Previous methods mitigate this issue by exploring image enhancement or object detection techniques with real low-light image datasets. However, the progress is impeded by the inherent difficulties about collecting and annotating low-light images. To address this challenge, we propose to boost low-light object detection with {zero-shot day-night domain adaptation}, which aims to generalize a detector from well-lit scenarios to low-light ones without requiring real low-light data. 
{Revisiting Retinex theory in the low-level vision,} we first design a reflectance representation learning module to learn Retinex-based illumination invariance in images with a carefully designed illumination invariance reinforcement strategy. Next, an interchange-redecomposition-coherence procedure is introduced to improve over the vanilla Retinex image decomposition process by performing two sequential image decompositions and introducing a redecomposition cohering loss. Extensive experiments on {ExDark, DARK FACE, and CODaN} datasets show strong low-light generalizability of our method. Our code is available at \small{\url{https://github.com/ZPDu/DAI-Net}}.
\end{abstract}

\vspace{-5px}    
\section{Introduction}
\label{sec:intro}

\begin{figure}

 \centering
\includegraphics[width=\linewidth]{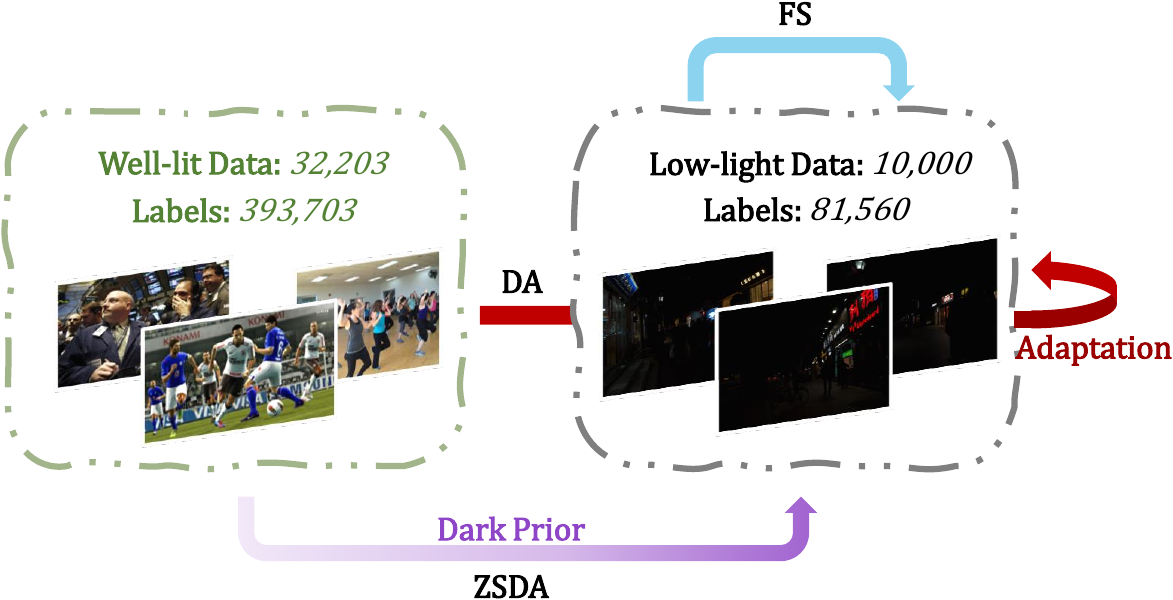}

\caption{ Left and right are respective well-lit and low-light datasets. FS: fully-supervised learning directly using low-light images and labels; DA: conventional domain adaptation with the access to low-light target domain data (and labels); ZSDA: {zero-shot day-night domain adaptation without the access of domain data but only knowing the dark scenario as a prior}.
}
	\vspace{-0.1in}
\label{fig:dataset}
\end{figure}

Object detection, aiming at identifying and localizing objects in an image, is a fundamental and well-investigated task in computer vision. Advanced detectors have achieved remarkable progress thanks to large-scale datasets such as COCO~\cite{lin2014eccv} and {Open Images~\cite{kuznetsova2020open}} for general object detection as well as WIDER FACE~\cite{yang2016cvpr} for human face detection~\cite{deng2020cvpr,guo2022iclr,liu2022mogface,liu2023damofd}. However, these methods encounter severe performance degradation on low-light images resulting from dark environments, inadequate lighting, and exposure time~\cite{wu2023cvpr}.
%as there are few low-light data in normal object detection benchmarks. 
Issues such as low visibility, color distortion, and noises arise in low-light images and impede the accuracy of object detectors. To tackle this challenge, image enhancement methods are normally investigated to enhance the visibility of the scene under low light~\cite{guo2020cvpr,ma2022cvpr,jin2022eccv,chen2018cvpr,xu2022cvpr,xu2020cvpr,wu2023cvpr,cai2023iccv}. Object detectors can benefit either through retraining or testing with {light-enhanced} images. Another direction is that detectors trained from well-lit images are fine-tuned on low-light images~\cite{wang2022acmmm,wang2021cvpr,wang2022tpami,sasagawa2020eccv}. 

Whilst existing image enhancement methods appear to be effective, they rely on a significant amount of low-light images collected from the real world. Many of them have to be trained with paired low-light and well-lit images {\cite{wei2018bmvc,wu2022cvpr}}. On the other hand, for dark object detection methods, the utilization of low-light images is also necessary~\cite{cui2021iccv,wang2021cvpr}. 
In contrast to the well-lit data, low-light images in established benchmarks~\cite{loh2019cviu,yang2020tip} are considerably sparser in amount and limited in terms of collecting scenarios. At the same time, annotating bounding boxes in low-light images is also undesirable due to low visibility. For instance, there are 32,203 images and 393,703 labelled faces in WIDER FACE, but only 10,000 images and 81,560 faces in DARK FACE. These difficulties in collecting and annotating low-light data hinder the development of low-light image enhancement and object detection.

To circumvent the requirement for object detection in low-light scenarios, we propose to work in a {zero-shot day-night domain adaptation setting}~\cite{lengyel2021iccv,luo2023iccv} as shown in Fig.~\ref{fig:dataset}, where object detectors are trained in the source domain with only well-lit images available and are evaluated in the low-light target domain with no images provided. 
The divergence between the well-lit source domain and the low-light target domain stems from {the illumination variation and the corruptions it brings}~\cite{cai2023iccv}.  
It is essential to emphasize that we are aware of the low-light scenario as our generalization target in this problem.

In this paper, we introduce a novel DArk-Illuminated Network, dubbed as DAI-Net, for {low-light object detection}. 
Given well-lit images from the source domain, we first utilize a physics-inspired low-illumination synthesis pipeline~\cite{cui2021iccv} to generate synthesized low-light images and form pairs with original well-lit images. 
{Next, we revisit the Retinex theory~\cite{land1977retinex} in low-level vision for the high-level detection task by} decomposing low-light images into domain-invariant (\ie image reflectance) and domain-specific (\ie image illumination) information, where only the former should be retained to learn a generalizable detector. To this end, we build our framework upon an established object detection pipeline (\eg DSFD~\cite{li2019cvpr}, YOLOv3~\cite{redmon2018yolov3}) and propose a reflectance representation learning module as an additional decoder.  This module decodes reflectance-related {illumination-invariant} information from well-lit images and synthetic low-light images. It is optimized with pseudo ground truth produced by a pre-trained Retinex decomposition network (\eg RetinexNet~\cite{wei2018bmvc}) and is reinforced with a specifically designed illumination invariance reinforcement strategy. Next, we design an interchange-redecomposition-coherence procedure to improve the Retinex-based image decomposition process. It performs two times image decomposition sequentially by interchanging the decomposed well-lit/low-light reflectance for reconstructing the images and then redecomposing them.
A redecomposition cohering loss is introduced to promote consistency between the produced reflectances in the two decompositions, so that the learned reflectance representation is stable and accurate.

In summary, to reduce the need for real-world low-light data, we propose to boost low-light object detection with {zero-shot day-night domain adaptation}. The contribution of this paper is threefold:

\begin{itemize}
    \item We introduce a reflectance representation learning module additional to an established object detector to enhance its illumination-invariance. Specifically, an illumination invariance reinforcement strategy is designed to strengthen the learning process.
    \item We propose an interchange-redecomposition-coherence procedure to improve the vanilla image decomposition process. A redecomposition cohering loss is introduced to maintain the consistency between the decomposition reflectances in sequence. 
    \item We conduct extensive experiments on ExDark for general object detection, DARK FACE for face detection, and {CODaN for image classification}. Experimental results show that our method outperforms the state-of-the-art in multiple settings by large margins. 
\end{itemize}
\section{Related Works}

\subsection{Object Detection}

Mainstream object detectors can be roughly classified into two categories: single-stage detectors, \eg SSD~\cite{liu2016eccv}, YOLO~\cite{redmon2018yolov3} and FCOS~\cite{tian2019iccv}; and two-stage detectors, \eg Faster R-CNN~\cite{ren2015nips} and R-FCN~\cite{dai2016nips}. Single-stage detectors aim to directly predict object bounding boxes and class labels in a single step. Two-stage detectors first generate a set of region proposals and then perform classification and bounding box regression to refine the proposals. {Recently, based on the transformer structure, DETR~\cite{carion2020eccv} and many DETR-based approaches~\cite{li2022cvpr,liu2022iclr,meng2021iccv,zhu2021iclr} are designed to further enhance general object detection. }

In contrast, face detectors~\cite{hu2017cvpr,najibi2017iccv,tang2018eccv,li2019cvpr,deng2020cvpr,guo2022iclr,liu2019point} are specialized detectors designed to detect human faces. 
Derived from general object detection, most face detection methods apply single-stage detectors to generate bounding boxes and corresponding classification scores simultaneously~\cite{li2019cvpr,deng2020cvpr,guo2022iclr,najibi2017iccv,tang2018eccv}. Many works focus on developing the model architectures~\cite{li2019cvpr,deng2020cvpr,guo2022iclr,liu2020cvpr}, anchor sampling/matching schemes~\cite{zhang2017iccv,ming2019cvpr,tang2018eccv}, as well as feature enhancement techniques~\cite{hu2017cvpr,lin2017cvpr,najibi2017iccv}. 

The recent prosperity of object detection should also be accredited to the availability of large-scale datasets. These datasets, such as COCO~\cite{lin2014eccv},  {Open Images~\cite{kuznetsova2020open}} {for general object detection}, and WIDER FACE~\cite{yang2016cvpr}, {FDDB~\cite{jain2010fddb} for face detection}, allowing researchers with ample annotated examples to train and assess their detection models.

\begin{figure*}

\centering
\includegraphics[width=\textwidth]{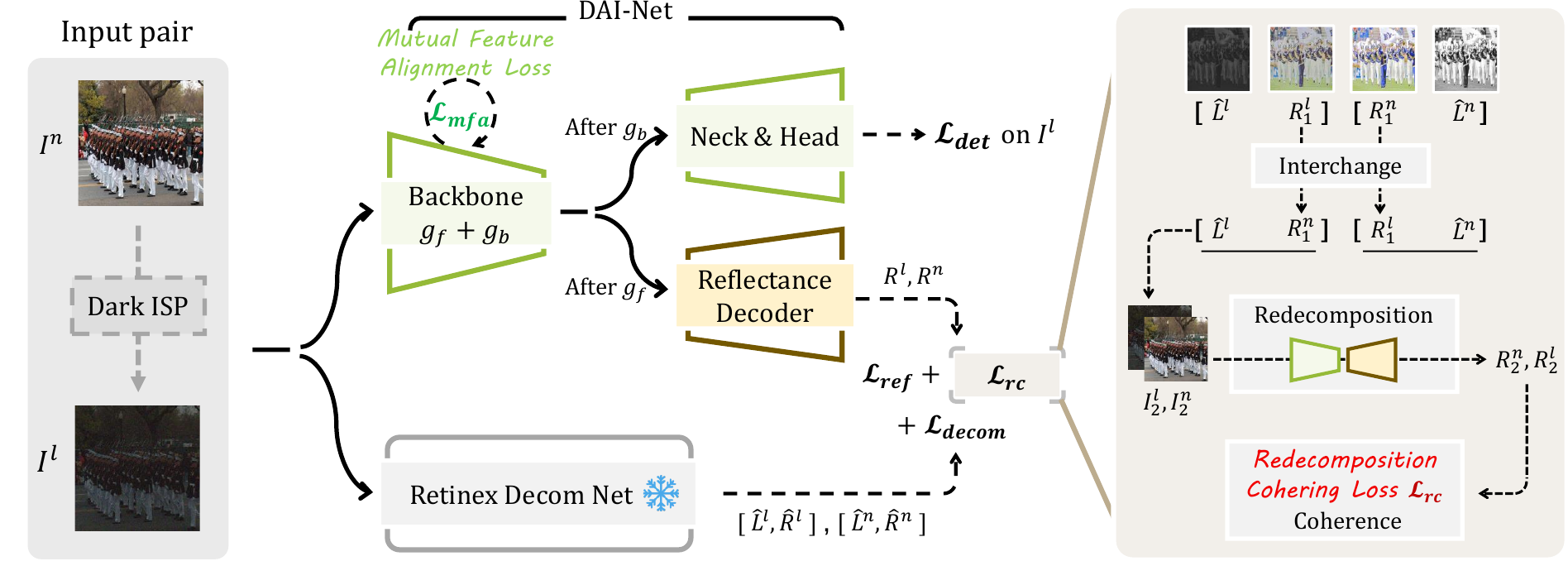}

\caption{ The structure overview of our method:  
we input a well-lit image and its corresponding synthesized low-light image as a pair into our framework. 
{The pretrained Retinex decomposition net, shown as the gray block on the bottom, is frozen and only used during training to infer reflectance and illumination pseudo ground truth $\hat{R}^l,\hat{R}^n,\hat{L}^l,\hat{L}^n$ to supervise the reflectance decoder (Sec.~\ref{sec:ref}). Specifically, illumination pseudo ground truth and the first-round reflectance predictions $R_1^l,R_1^n$ are forwarded into the proposed interchange-redecomposition-coherence procedure on the right block for {reconstructing and redecomposing second-round reflectance predictions $R_2^l,R_2^n$, as well as calculating redecomposition cohering loss $\mathcal{L}_{rc}$} (Sec.~\ref{sec:redecomp}). The inference is the same as the original detector that our method is built on, represented as green blocks. }
}
\label{fig:overview}
\vspace{-0.2in}
\end{figure*}
\subsection{Low-Light Images}
\label{sec:retinex}
\textbf{Low-Light Image Enhancement.}  Low-light image enhancement has been largely explored owing to the advent of deep learning.
%~\cite{guo2020cvpr,ma2022cvpr,jin2022eccv,lore2017pr,chen2018cvpr,i2023iclr,xu2022cvpr,xu2020cvpr,wu2023cvpr,cai2023iccv,yi2023iccv}. 
For instance, Guo~\etal\cite{guo2020cvpr} propose zero-DCE to estimate image-specific {light-enhancement curves} without reference images. Many works~\cite{cai2017iccv,wei2018bmvc,yang2021tip,wang2019cvpr,liu2021cvpr,wu2022cvpr,jin2022eccv,cai2023iccv} base themselves on the Retinex theory~\cite{land1977retinex}, {which assumes that an image can be decomposed into reflectance and illumination and can be reconstructed via element-wise multiplication of the latter two. Images can thus be enhanced by two major ways: using the reflectance as enhanced image~\cite{guo2016tip,wang2019cvpr,ma2022cvpr,cai2023iccv}, or reconstructing an enhanced image with adjusted illumination~\cite{wei2018bmvc,cai2017iccv,jin2022eccv,wu2022cvpr,liu2021cvpr,yi2023iccv}. For instance, Wei~\etal\cite{wei2018bmvc} conduct light enhancement and denoising based on reflectance and illumination maps decomposed from a Retinex decomposition model.  %\zp{for lighted illumination and denoised reflectance}. 
Liu~\etal\cite{liu2021cvpr} unroll the optimizing process of Retinex-based model and find the desired network structure with a cooperative bilevel search strategy. Cai~\etal\cite{cai2023iccv} simultaneously model corruptions in reflectance and illumination through a transformer structure. }

\noindent\textbf{Low-light Object Detection.} Low-light object detection methods can be mainly categorized into three groups: detection-by-enhancement~\cite{guo2020cvpr,ma2022cvpr,qin2022accv}, enhancement-for-detection~\cite{sun2022nips,wang2022acmmm,ma2022acmmm} and low-light detector learning strategies~\cite{sasagawa2020eccv,cui2021iccv,wang2021cvpr,hashmi2023iccv}. In detection-by-enhancement, low-light enhancement methods are employed to generate illuminated images before detection~\cite{guo2020cvpr,ma2022cvpr,qin2022accv}. In enhancement-for-detection, representative works are \cite{sun2022nips,hashmi2023iccv} which propose image restoration training pipelines for improving object detection performance. For low-light detector learning strategies, multi-model merging~\cite{sasagawa2020eccv}, multi-task auto encoding transformation~\cite{cui2021iccv} and unsupervised domain adaptation framework~\cite{wang2021cvpr} are introduced. 
In addition, there also have been efforts to construct low-light datasets, such as Nightowls~\cite{neumann2019accv}, ExDark~\cite{loh2019cviu}, DARK FACE~\cite{yang2020tip}, NOD~\cite{morawski2021bmvc}. However, these datasets are not as adequate as well-lit ones. Our method falls into the third category and addresses the concern for low-light data shortage in a zero-shot day-night domain adaptation setting.
% in a new zero-shot dark domain generalization setting.  

\noindent\textbf{Dark Domain Learning.} Many perception tasks such as object segmentation and detection in dark scenarios have been investigated under domain transfer learning, which can be broadly categorized into domain adaptation and generalization. Domain adaptation (DA) methods train a model on well-lit source domain and adapt it to the accessible low-light target domain data. Representative attempts include synthesizing low-light images~\cite{arruda2019ijcnn,mukherjee2022bdat,romera2019iv,sakaridis2019iccv,wu2021cvpr,cui2021iccv,gao2022cvpr},  {self-supervised learning for aligning well-lit and low-light domain distributions~\cite{wang2021cvpr,wang2022tpami,wang2022acmmm},  merging components learned in two domains}~\cite{sasagawa2020eccv,vankadari2020eccv}, and many other multi-stage strategies~\cite{wang2021cvpr,ye2022cvpr,kennerley2023cvpr,deng2022cvpr,sakaridis2019iccv,dai2018dark}.  In contrast, domain generalization (DG) differs from DA by generalizing to unseen domains without knowing the knowledge of the target domain~\cite{fan2023iclr,vidit2023cvpr,reddy2022eccv,du2023domain}. The DG methods aim to provide general solutions to a broad range of potential target domains instead of focusing on one single domain. For low-light {scenarios}, a setting coined as zero-shot day-night domain adaptation (ZSDA) is introduced in~\cite{lengyel2021iccv,luo2023iccv}, which can actually be regarded as a special case of DA: real low-light data are inaccessible but the target domain is known to be the low-light scenario. Our work falls into the realm of ZSDA and we are the first to focus on ZSDA detection.
{A prior work,~\cite{wang2021cvpr}, also achieves low-light detection but belongs to unsupervised DA, which leverages unlabeled real low-light data, different from ZSDA.} 

\section{Method}
\label{sec:method}

\subsection{Overview}

We train an object detector on well-lit images and generalize it to low-light images. 
{Following~\cite{lengyel2021iccv,luo2023iccv}, this is a zero-shot day-night domain adaptation (ZSDA) problem from well-lit source domain to low-light target domain. The primary factors leading to the disparity between the two domains include illumination and the corruptions brought by illumination degradation~\cite{cai2023iccv}.}

To resolve this gap, we focus on acquiring domain-invariant representations {from the source domain} that can transcend the target domain. {This is specifically interpreted as learning illumination-invariant information, \ie the reflectance representation, in the ZSDA setting}. 
We illustrate the proposed framework in Fig.~\ref{fig:overview}. During training, the framework consists of the proposed DArk-Illuminated Network (green and yellow blocks in Fig.~\ref{fig:overview}) and a pretrained Retinex decomposition net (gray block in Fig.~\ref{fig:overview}). The input to both networks is the same, which is a pair of a well-lit source domain image and its low-light counterpart synthesized by a physics-inspired low-light degradation synthesis pipeline~\cite{cui2021iccv}, termed as Dark ISP in Fig.~\ref{fig:overview}. {Dark ISP is a non-deep method and is only used to create paired input for training our framework.} It is noteworthy that training the detector directly on synthetic low-light images leads to much worse results than training on well-lit images (see Sec.~\ref{sec:ab}). The proposed DArk-Illuminated Network (DAI-Net) is built on an established object detector with an additional decoder for reflectance representation learning (Sec.~\ref{sec:ref}). A pretrained Retinex decomposition net is leveraged to provide pseudo ground truth for this decoder. An interchange-redecomposition-coherence procedure (Sec.~\ref{sec:redecomp}) is introduced to further strengthen the {reflectance representation learning.}

\subsection{Learning Retinex-based Reflectance Representation}
\label{sec:ref}

We first revisit the Retinex theory~\cite{land1977retinex}, 
an image $I$ can be decomposed into the reflectance $R $ and illumination $L$, $I = R \cdot L$. The theory posits that the visibility of an image is influenced by illumination, while the reflectance remains unchanged. Sometimes the corruptions caused by the illumination variation are also jointly modeled, whilst we simply term them as part of illumination for convenience (Sec.~\ref{sec:generalizability}). For object detection with ZSDA, we consider the reflectance as the illumination-invariant counterpart, the acquisition of reflectance knowledge becomes instrumental in achieving an illumination-invariant detector. To this end, we design a reflectance representation learning module to enhance the detector's resilience to low-light effect. 

{Considering image decomposition is rather a low-level vision task},
we split the detector backbone {at the second Conv layer, and denote this frontal part of the backbone as $g_{f}$ and the whole backbone as $g_b$}. The output of $g_{f}$, \ie feature $F$, encodes low-level information extracted {by shallow layers}, which is suitable for decoding the reflectance.
We branch off the reflectance decoder after $g_f$, shown as the {yellow block} in Fig.~\ref{fig:overview}. {The decoder {is a light module consisting} of two Conv+ReLU layers.} Since the detection head shares $g_f$ with the reflectance decoder, the extracted illumination-invariant features also benefit the object detection task. Notice that we find learning image decomposition on the fly with object detection is unstable and could fail in some scenarios. We leverage a pretrained Retinex decomposition net~(see Sec.~\ref{sec:train} for details) to generate reflectance and illumination {pseudo} ground truth {\{$\hat{R},\hat{L}\}$} to supervise reflectance decoder in a more stable manner. 

\textit{Illumination invariance reinforcement.} We further introduce an illumination invariance reinforcement scheme to reinforce the illumination invariance of the detector {from feature level}. {Illumination discrepancy between the paired well-lit and synthesized low-light input images can lead to different feature distributions. But the two images are in essence with the same semantic information. Since our target is to learn illumination invariant representation, we require the {output features $F$ from $g_f$} forwarded into the reflectance decoder to be {closely aligned} between well-lit and low-light images.} We explicitly match the well-lit and low-light features $F^n$ and $F^l$ {extracted from $g_f$} by designing a {mutual feature alignment loss} as follows:
\begin{equation}
\label{eq:mc}
    \mathcal{L}_{mfa} = \mathcal{KL}(F^n||F^l) + \mathcal{KL}(F^l||F^n),
\end{equation}
where $\mathcal{KL}(\cdot||\cdot)$ refers to KL-Divergence, $F^l$ and $F^n$ are flattened and spatially averaged features from $g_{f}$ \wrt well-lit and synthesized low-light images, respectively. 

\subsection{Interchange-Redecomposition-Coherence}
\label{sec:redecomp}

To further enhance the reflectance learning, we aim at designing a stronger image decomposition process. {Given a pair of low-light image $I^l$ and well-lit image $I^n$, a {typical} Retinex-based image decomposition process  ~\cite{yang2021tip,wang2019cvpr,liu2021cvpr,wu2022cvpr,jin2022eccv,cai2023iccv} decomposes them into corresponding reflectance and illumination, \ie low-light reflectance $R^l_1$ and illumination $L^l$ for $I^l$ and well-lit reflectance $R^n_1$ and illumination  $L^n$ for $I^n$. Both reflectance $R^l_1$ and $R^n_1$ ({they should be ideally the same}) should be interchangeable with each other to reconstruct $I^n,I^l$ when combined with the corresponding illumination map $L^n,L^l$.} 
Thanks to this interchangeability,
we can add a constraint to strengthen the image decomposition and reflectance representation learning. {One intuitive approach is to impose a penalty on the images that are reconstructed using the interchanged reflectance once they deviate from the original inputs, as proposed in~\cite{wei2018bmvc}.} This is considered as a {vanilla penalty} loss constraint. To fully harness the information produced in image decomposition, we propose an interchange-redecomposition-coherence procedure, which is depicted in the right of Fig.~\ref{fig:overview}. First, we interchange the reflectance between {well-lit and low-light images} and reconstruct the images as $I_2^l=R^n_1\cdot L^l, I_2^n=R^l_1\cdot L^n$. 
{The reconstructed images can be decomposed into a second round. As the emphasis of our DAI-Net is to learn the illumination-invariant part of the image (the reflectance), we decompose reflectances $R^n_2,R^l_2$ from $I_2^n,I_2^l$ using the same reflectance decoding branch in DAI-Net as in the first round and request them to be coherent to corresponding reflectances $R^l_1,R^n_1$ decomposed in the first round. We, therefore, introduce the {redecomposition} cohering loss:}
\begin{equation}
\label{eq:red}
    \mathcal{L}_{rc} = ||R_1^n-R_2^l||_1 + ||R_1^l-R_2^n||_1 
\end{equation}
Compared with the vanilla penalty loss, $\mathcal{L}_{rc}$ involves a redecomposition process and takes full advantage of the interchangeability of reflectance. 

\subsection{Network Training}
\label{sec:train}

{\textbf{Retinex decomposition net.} 
{The Retinex decomposition net is an off-the-shelf network that can be based on any image decomposition networks (\eg~\cite{wei2018bmvc,cai2023iccv}).} 
We can train this network {from scratch  using our paired input (default setting), or directly load pretrained weights that are publicly available.}
Afterwards, we    
freeze it to only infer the reflectance and illumination of the input as pseudo ground truth {\{$\hat{R},\hat{L}\}$} during the learning of the {DAI-Net}. {We choose a basic structure~\cite{wei2018bmvc} by default, as perfect pseudo labels are not a necessity for enabling good detection performance.}

\noindent\textbf{DAI-Net.}  
{The DAI-Net consists of a detection branch and a reflectance decoding branch. For the former, we utilize the detection losses of the selected detector, denoted by $\mathcal{L}_{det}$. }{While the objective functions of reflectance decoding branch are summarized into three parts. The first part consists of the two proposed losses $\mathcal{L}_{mfa},\mathcal{L}_{rc}$ (Eq.~\ref{eq:mc} and \ref{eq:red}). The other two are: }

\noindent{\textit{Reflectance learning loss.} We supervise the reflectance decoder output $R$ by the pseudo ground truth $\hat{R}$ through a reflectance learning loss $\mathcal{L}_{ref}=\text{MAE}(R,\hat{R})+(1-\text{SSIM}(R,\hat{R}))$ where MAE is the Mean Absolute Error and SSIM is the Structural Similarity Index Measure.} $(R,\hat{R})$ is realized as $(R^l,\hat{R}^l)$ or $(R^n,\hat{R}^n)$ in practice.}

\noindent{\textit{Image decomposition loss}. We further strengthen the reflectance learning through image decomposition loss. We apply a representative decomposition loss $\mathcal{L}_{decom}$, as in~\cite{wei2018bmvc}, upon $\hat{L}$ and $R$. Specifically, the loss is an integration of image reconstruction loss $\mathcal{L}_{recon}$, invariant reflectance loss $\mathcal{L}_{ir}$, and illumination smoothness loss $\mathcal{L}_{smooth}$. $\mathcal{L}_{recon}$ is to let $R \cdot \hat{L}$ reconstruct the input image $I$, specifically, $R^l \cdot \hat{L}^l$ for $I^l$ and $R^n \cdot \hat{L}^n$ for $I^n$; $\mathcal{L}_{smooth}$ and $\mathcal{L}_{ir}$ are computed between the paired input. $\mathcal{L}_{ir}=MSE(R^l,R^n)+(1-SSIM(R^l,R^n))$ enforces the predicted well-lit reflectance and low-light reflectance to be the same, in the form of a combination of Mean Squared Error and Structure Similarity Index Measure~\cite{wang2004ssim}. Therefore, we have}  
\begin{equation}\mathcal{L}_{decom}=\mathcal{L}_{recon}+\lambda_{smooth}\mathcal{L}_{smooth}+\lambda_{ir}\mathcal{L}_{ir}, 
\end{equation}
\noindent where $\lambda_{smooth}, \lambda_{ir}$ represents loss weight for corresponding loss, respectively. 

The total objective function of DAI-Net can be formulated as:

% \vspace{-0.1in}
\begin{equation}
\label{eq:overall}
    {\mathcal{L} = \mathcal{L}_{det}  + \lambda_{mfa} \mathcal{L}_{mfa} + \lambda_{rc}\mathcal{L}_{rc} + \mathcal{L}_{ref} + \mathcal{L}_{decom}}
\end{equation}
where $\lambda_{mfa}$ refers to loss weight for mutual feature alignment loss in {Eq.~\ref{eq:mc}} and $\lambda_{rc}$ refers to loss weight for redecomposition cohering loss in Eq.~\ref{eq:red}.

\section{Experiments}

% \subsection{Experimental Settings}

In this section, we first provide settings and results on three tasks: {face detection (Sec.~\ref{subsec:darkface}), object detection (Sec.~\ref{subsec:objectdetection}) and image classification (Sec.~\ref{sec:generalizability}) in darkness.} The first two tasks validate the proposed detection method, while the last one shows the generalizability of our approach. 
Furthermore, we ablate the proposed components and conduct corresponding analyses. 

\begin{table}[!t]

          \centering
 \footnotesize

			\begin{tabular}{ll?c}
				\toprule[1.5pt]
				\textbf{Category} & \textbf{Method} & \textbf{mAP(\%)}  \\
				\midrule[1.5pt]
                \multicolumn{3}{l}{\textbf{WIDER FACE$\rightarrow$DARK FACE test set using DSFD}}\\
                \midrule
                \multirow{7}{*}{\textbf{Generalization}} & Faster-RCNN~\cite{ren2015nips} & 1.7\\
                & SSH~\cite{najibi2017iccv} & 6.9\\
                & RetinaFace~\cite{deng2020cvpr}& 8.6\\
                & SRN~\cite{chi2019aaai} & 9.0\\
                & SFA~\cite{luo2019access} & 9.3\\
                & ParamidBox~\cite{tang2018eccv} & 12.5\\
                & SmallHardFace~\cite{zhang2020wacv} & 16.1\\
                & DSFD~\cite{li2019cvpr}& 16.1\\

                \midrule
                \multirow{3}{*}{\textbf{Zero-shot Adaptation}} & CIConv~\cite{lengyel2021iccv}& 18.4\\
                 & Sim-MinMax~\cite{luo2023iccv} & 25.7\\
                   & \textbf{DAI-Net} & \textbf{28.0}\\
                % \midrule
                %    \textbf{Unsupervised Adaptation} & HLA-Face~\cite{wang2021cvpr} & 44.4\\
                \midrule
                \multirow{2}{*}{\textbf{Fully Supervised}} & Fine-tuned DSFD~\cite{wang2021cvpr}& 46.0\\
                 & \textbf{Fine-tuned DAI-Net}& \textbf{52.9}\\
                \midrule[1.5pt]
                \multicolumn{3}{l}{\textbf{COCO$\rightarrow$DARK FACE validation set~\cite{cui2021iccv} using YOLOv3}} \\
                \midrule
                 \multirow{7}{*}{\textbf{Pretrained and Tuning}} & YOLO$_N$ & 48.3\\
                 & YOLO$_N$+MBLLEN~\cite{lv2018bmvc} & 51.6\\
                 & YOLO$_N$+KIND~\cite{zhang2019acmmm} & 51.6\\
                 & YOLO$_N$+Zero-DCE~\cite{guo2020cvpr} & 54.2\\
                % \midrule
                & YOLO$_L$&54.0\\
                & MAET~\cite{cui2021iccv} & 55.8\\
                % \midrule
                & \textbf{DAI-Net} & \textbf{57.0}\\
                 \bottomrule[1.5pt]
			\end{tabular}
    \vspace{-2px}
	\caption{Comparison with state of the art on DARK FACE test and validation sets. }
 	\label{table:darkface}
  \vspace{-0.2in}
\end{table}

\begin{table*}[!t]
	% \vspace{-0.3in}
	\small
 \setlength{\tabcolsep}{1.8mm}
  \centering

		% \begin{sc}
			\begin{tabular}{c?cccccccccccc?c}
				\toprule[1.5pt]
				 \textbf{Method} & \textbf{Bicycle} & \textbf{Boat} & \textbf{Bottle} & \textbf{Bus} & \textbf{Car} & \textbf{Cat} & \textbf{Chair} & \textbf{Cup} & \textbf{Dog} & \textbf{Motorbike} & \textbf{People} &\textbf{Table} & \textbf{Total}\\
		
                \midrule[1.5pt]
                YOLO$_N$ & 71.8 & 64.5 & 63.9 & 81.6 & 76.8 & 55.4 & 49.7 & 56.8 & 63.8 & 61.8 & 65.7 & 40.5 & 62.7\\
                +KinD~\cite{zhang2019acmmm} & 73.4 & 68.1 & 65.5 & 86.2 & 78.3 & 63.0 & 56.9 & 62.7& 68.2 & 67.1 & 69.6 & 48.2 & 67.3\\
                +Zero-DCE~\cite{guo2020cvpr} & 79.5 & 71.3 & 70.4 & 89.0 & 80.7 & 68.4 & 65.7 & 68.6 & 75.4 & 67.2 & 76.2 & 51.1 & 72.0\\
                \midrule
                YOLO$_L$ & 78.2 & 70.8 & 72.3 & 88.1 & 80.7 & 67.9 & 62.4 & 70.5 & 74.8 & 69.4 & 75.8 & 50.9 & 71.6\\
                MAET~\cite{cui2021iccv} & 81.3& 71.6 & 74.5 & 89.7 & 82.1 & 69.5 & 65.5 & 72.6 & 75.4 & 72.7 & 77.4 & 53.3 & 74.0\\
                \midrule
                \textbf{Ours} & \textbf{83.8} & \textbf{75.8} & \textbf{75.1} & \textbf{94.2} & \textbf{84.1} & \textbf{74.9} & \textbf{73.1} & \textbf{79.2} & \textbf{82.2} & \textbf{76.4} & \textbf{80.7} & \textbf{59.8}& \textbf{78.3}\\
                 \bottomrule[1.5pt]
			\end{tabular}
 	\caption{Comparison with state of the art on ExDark.}
	\label{table:exdark}
	\vspace{-0.1in}
\end{table*}

\subsection{Face Detection in Darkness}
\label{subsec:darkface}

\textbf{Settings.} We conduct experiments by selecting HLA-Face~\cite{wang2021cvpr} and MAET~\cite{cui2021iccv} as comparable methods. They both reproduce and report the results of a number of methods in low-light settings. Based on their configurations, we select WIDER FACE~\cite{yang2016cvpr} and COCO 2017~\cite{lin2014eccv} as two separate well-lit source domains, which contain over 100K and 30K images, respectively. We test the trained models on the target domain DARK FACE containing around 10K images. These are denoted as WIDER FACE/COCO$\rightarrow$DARK FACE (from source to target). We consistently adopt the same base face detector DSFD~\cite{li2019cvpr}/YOLOv3~\cite{redmon2018yolov3} and plug DAI-Net on top of it. We train the network following the same training process outlined in~\cite{li2019cvpr,cui2021iccv}. 
% We pretrain Retinex decomposition net 
{We build Retinex decomposition net using RetinexNet~\cite{wei2018bmvc} and pretrain it}
as described in Sec.~\ref{sec:train} for 10 epochs with a learning rate of 1e-3. Loss weights $\lambda_{smooth}$ and $\lambda_{ir}$ are set to 0.5 and 0.01, respectively.  $\lambda_{rc}$ and $\lambda_{mfa}$ in Eq.~\ref{eq:overall} are 0.001 and 0.1. {The optimal weights for $\lambda_{smooth}$ and $\lambda_{ir}$ can be easily found around their default settings in~\cite{wei2018bmvc}. We mainly tune $\lambda_{rc}$ and $\lambda_{mfa}$ on the WIDER FACE validation set.} 

{We evaluate our method using the commonly used metric in object detection, \ie mean Average Precision(mAP). We compare our results with results presented in HLA-Face~\cite{wang2021cvpr} and MAET~\cite{cui2021iccv} in Table~\ref{table:darkface}. We use the same multi-scale scheme as in ~\cite{li2019cvpr,wang2021cvpr}  {to DAI-Net} for testing. }

\begin{table}[t]
    \centering

	% \vspace{-0.1in}

	    \centering

        {

			\begin{tabular}{c?c}
				\toprule[1.5px]
				\textbf{Method} & \textbf{Top-1 (\%)}\\
                \midrule[1.5px]
                MAET~\cite{cui2021iccv} & 56.48\\
                CIConv~\cite{lengyel2021iccv} & 60.32\\
                Sim-MinMax~\cite{luo2023iccv} & 65.87\\
                \midrule
                \textbf{Ours} & \textbf{68.44}\\

                 \bottomrule[1.5px]
			\end{tabular}

	}
 \vspace{-2px}
	\caption{Task generalization on CODaN.}
 \vspace{-10px}
 	\label{table:taskgen}
\end{table}

\noindent{\textbf{{WIDER FACE$\rightarrow$DARK FACE. }}We compare with the results in HLA-Face~\cite{wang2021cvpr} using DSFD as they do.} 
\noindent\underline{\textit{Generalization.}} In the first group of {Table~\ref{table:darkface}}, comparable methods are trained on the source domain and directly evaluated on the target domain. {DSFD achieves the best generalizability, which we base our framework on.} Our method further improves upon DSFD by a large margin. 

\noindent\underline{\textit{Zero-shot Adaptation.}} 
Under this setting, the methods are aware that the target domain is the dark scenario. We directly compare with ZSDA methods~\cite{lengyel2021iccv,luo2023iccv} by applying them to DSFD for detection. Ours achieves the best result, showing a strong dark domain generalizability. 

\noindent\underline{\textit{Fully Supervised.}} In this group, the trained models from the first group (\emph{Generalization}) are fine-tuned {in their respective object detection learning manners} on the training set of DARK FACE with ground truth labels. Our fine-tuned DAI-Net still outperforms other comparable methods by a large margin. Our method hence contributes to a new baseline for this dataset for future development.

\noindent\textbf{{COCO$\rightarrow$DARK FACE. }} We compare with MAET~\cite{cui2021iccv} using YOLOv3 {trained on COCO and finetuned the same way} on DARK FACE with labels.
The results in MAET~\cite{cui2021iccv} are reported on a specifically sampled validation set. For a direct comparison with it, we use its published train/val division to have 5500 training images {for finetuning DAI-Net} {pretrained on COCO} and 500 validation images to evaluate it. The result is shown in Table~\ref{table:darkface}, under the category of COCO$\rightarrow$DARK FACE. 
Our model performs clearly better than {MAET and other comparable methods reported in~\cite{cui2021iccv}}.

\begin{table}[!t]

 \setlength{\tabcolsep}{1.5mm}

 \footnotesize	
  \centering

			\begin{tabular}{c|ccc|cc|ccc?c}
				\toprule[1.5px]
				\multicolumn{9}{c?}{Method} & \multirow{3}{*}{mAP(\%)}  \\\cmidrule{1-9}
                \multirow{2}{*}{DISP} & \multicolumn{3}{c|}{{RD}} & \multicolumn{2}{c|}{DP} & \multicolumn{3}{c?}{$\mathcal{L}_{mfa}$} & \\\cmidrule{2-9}
                & R& L&R+L &{$\mathcal{L}_{p}$}&{$\mathcal{L}_{rc}$}& KL & L1& L2& \\
				\midrule[1.5px]

                -- & --& --& -- & -- & --& --& --& -- & 15.2 \\
                \checkmark & --& --& -- & --& --& --& --& --& 11.6\\
                \midrule
                \checkmark & \checkmark &-- &-- & -- & -- & --& --& --& 20.5\\
                \checkmark & -- & \checkmark & -- & -- & -- & --& --& --&16.5\\
                \checkmark & -- & -- & \checkmark & -- & -- & --& --& --&19.0\\
                \midrule
                \checkmark & \checkmark & --& --& \checkmark & -- & -- & --& --& 21.8\\
                \checkmark & \checkmark & -- & -- & -- & \checkmark & -- & --& --&  22.3\\
                \checkmark & \checkmark & -- & --& \checkmark & \checkmark & -- & --  & --& 22.0\\
                \midrule
                \checkmark & \checkmark &-- & --& --& \checkmark & \checkmark & -- & --& \textbf{23.5}\\
                \checkmark & \checkmark &-- & --& --& \checkmark & -- & \checkmark & --& 20.7\\
                \checkmark & \checkmark &-- & --& --& \checkmark & -- & -- & \checkmark& 21.1\\
                 \bottomrule[1.5px]
			\end{tabular}

        \vspace{-0.05in}

	\caption{Ablation Study of proposed components on DARK FACE.}
    \vspace{-0.15in}
	\label{table:ablation}

\end{table}

\subsection{Object Detection in Darkness}
\label{subsec:objectdetection}

\begin{figure*}[!t]

% \begin{overpic} 
% [width=\linewidth]
% {example-image-a}
% \end{overpic}
\centering
\includegraphics[width=0.9\textwidth]{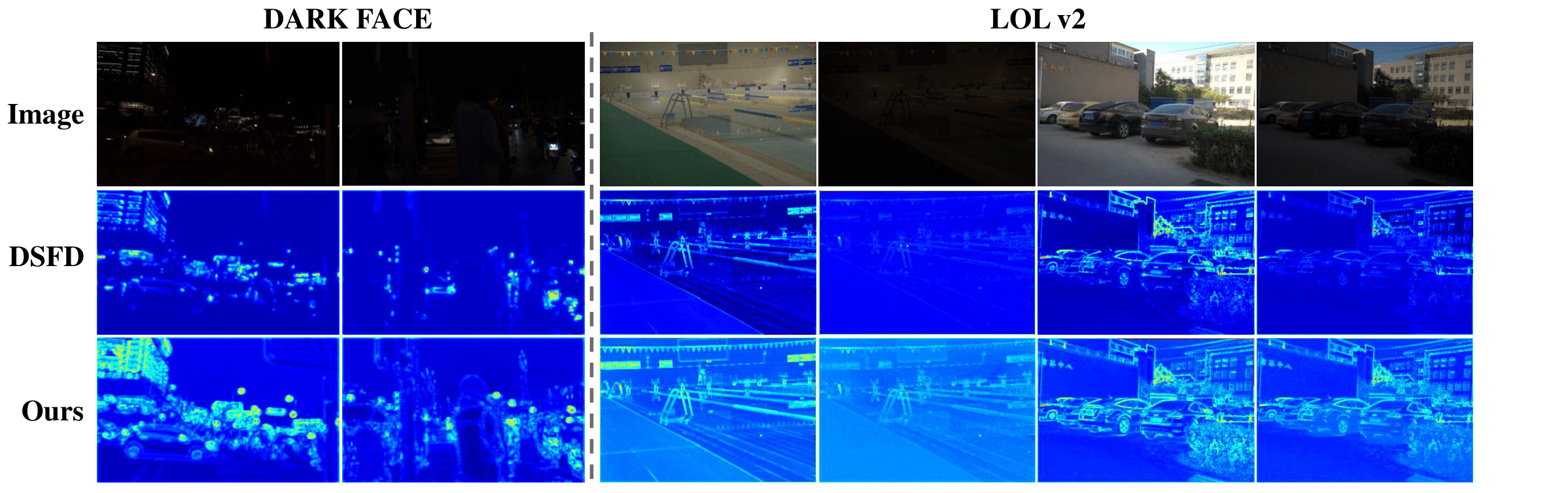}
\vspace{-0.1in}
\caption{ 
Feature visualization. The original image, backbone features from DSFD~\cite{li2019cvpr}, and our method are given in the three rows.
}
\vspace{-0.1in}
\label{fig:feat}

\end{figure*}

{\textbf{Settings.} Following MAET~\cite{cui2021iccv}, we use YOLOv3 as the object detector and take COCO as the source domain and ExDark~\cite{loh2019cviu} as the target domain. ExDark has around 7K dark images. The settings~{of hyperparameters} are consistent as in Sec.~\ref{subsec:darkface}.}

\noindent\textbf{Results. }In Table~\ref{table:exdark}, we compare our results to the state of the art. YOLO$_N$ and YOLO$_L$ refer to training {YOLOv3 based DAI-Net} on original well-lit images {of COCO} and synthesized low-light images {generated by {the same Dark ISP}~\cite{cui2021iccv}}, respectively. Our method shows superior performance on all object categories and the best overall result. For instance, our method enhances mAP on the difficult class \emph{Chair} by 7.6\%. 
This implies our proposed {modules in DAI-Net} can be easily plugged into general object detectors to boost their performance in dark object detection.

\subsection{Generalization to Dark Image Classification}
\label{sec:generalizability}

We show the generalizability of our approach beyond the dark object detection task by applying our method to the dark image classification task. This allows us to directly compare with the reported results in ZSDA methods~\cite{cai2023iccv,lengyel2021iccv}.

\noindent\textbf{Settings.} Following Sim-MinMax~\cite{cai2023iccv}, we conduct experiments on a nighttime image classification dataset CODaN~\cite{lengyel2021iccv}. CODaN {covers 10000 well-lit training samples, 2500 well-lit test samples for validation}, and 2500 low-light test samples.
We use the same ResNet-18~\cite{he2016deep} backbone.   
The hyperparameters of our method remain as in Sec.~\ref{subsec:darkface}.

\noindent\textbf{Results.} The results on CODaN are shown in Table~\ref{table:taskgen}. Our results surpass the previous best by a large margin. This demonstrates ours can be effectively applied to other tasks.

\subsection{Ablation Study} 
\label{sec:ab}
We conduct ablation studies on {WIDER FACE $\rightarrow$ DARK FACE }~{for face detection in darkness} to validate the effectiveness of proposed components. The multi-scale testing strategy is not used for evaluation efficiency. {More experiments are provided in supplementary materials.}

\begin{figure}[!t]

\centering
\includegraphics[width=0.9\linewidth]{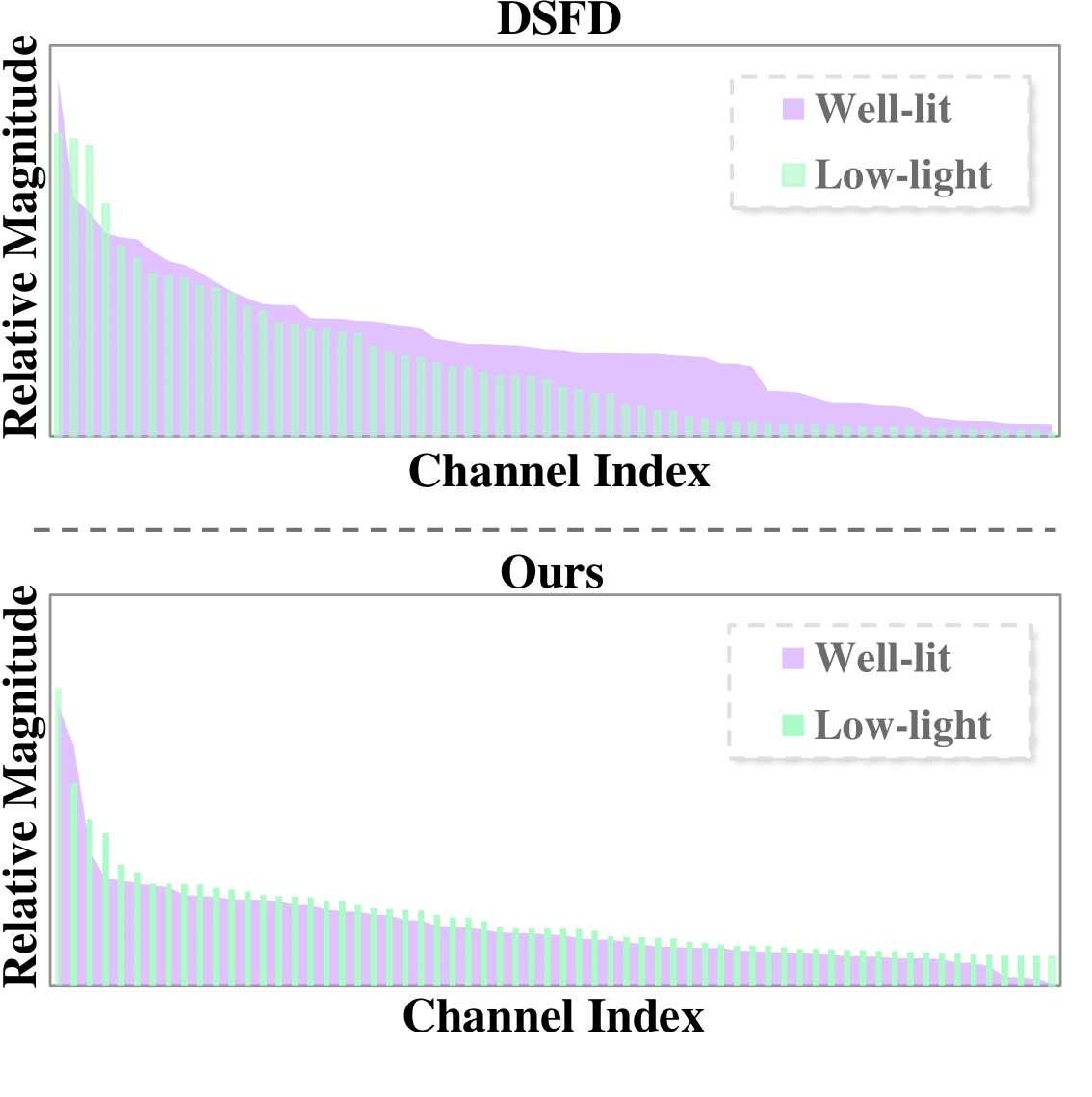}
\vspace{-0.2in}
\caption{ 
Mean magnitude of feature channels before and after applying our method. {We use discrete and continuous curves for the two distributions respectively only for the visualization purpose.}
}
\vspace{-0.2in}
\label{fig:featstat}

\end{figure}

\begin{figure*}[!t]

\centering
\includegraphics[width=0.98\textwidth]{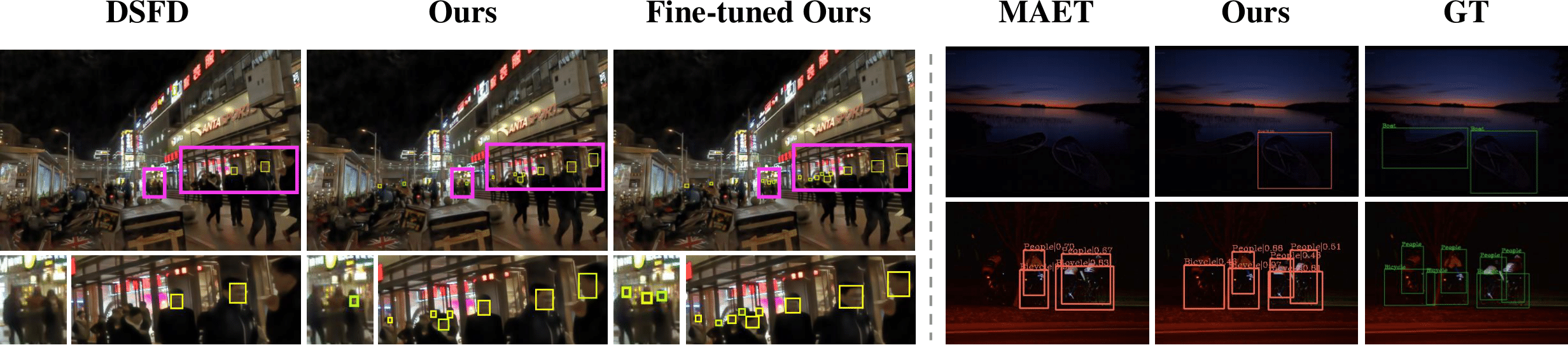}

\caption{ Left: Qualitative examples of {DSFD~\cite{li2019cvpr}}, ours and fine-tuned ours (see Sec.~\ref{subsec:darkface}) on DARK FACE. {DARK FACE images} are enhanced only for visualization. Right: Qualitative examples of MAET{~\cite{cui2021iccv}}, ours {and the ground truth} on ExDark (zoom in for details).
}

\label{fig:result}
\vspace{-3px}
\end{figure*}

\noindent\textbf{Baseline.} Our baseline model is the vanilla {DSFD} model~\cite{li2019cvpr} {where we do not add any component proposed in DAI-Net}. We report its result in the first row of Table~\ref{table:ablation}. We apply the Dark ISP~~\cite{cui2021iccv} to synthesize low-light images, which is denoted as the DISP column in Table~\ref{table:ablation}. Directly training on synthesized images {degrades the precision of {DSFD} from 15.2\% to 11.6\%}, implying that synthesized {low-light} images are {not} of the same attributes as real low-light data.

\noindent\textbf{Reflectance Decoding. } We add the reflectance decoding branch to the baseline. This is denoted as R under column RD of Table~\ref{table:ablation}. The result is shown in the third row in Table~\ref{table:ablation}. The performance is largely boosted { from 15.2\% to 20.5\%}. Adding the detector with an auxiliary head to generate reflectance helps to enhance the illumination-invariant information in the detector, and thus leads to more robust detection under various illuminations.

\noindent\underline{\textit{Learning Target.}} Based on Retinex theory, we identify the reflectance as an illumination-invariant target. To validate this, we try other variants such as adding auxiliary branches to DAI-Net for decoding illumination or decoding both reflectance and illumination, denoted by respectively adding L and R+L under column RD in Table~\ref{table:ablation}. Both variants are inferior to our original version, {which decodes only the reflectance.}

\noindent\textbf{Decomposition Processes.} We investigate the effectiveness of the interchange-redecomposition-coherence procedure by comparing redecomposition cohering loss with the vanilla {penalty loss~\cite{wei2018bmvc} discussed in Sec.~\ref{sec:redecomp}}. In Table~\ref{table:ablation}, we present the results among the sixth to eighth rows.

\noindent\underline{\textit{Image Decomposition Loss.}} We first apply the representative {penalty loss ($\mathcal{L}_{p}$, {see Sec.~\ref{sec:redecomp}})}{ to replace $\mathcal{L}_{rc}$ in DAI-Net}. Table~\ref{table:ablation} shows that $\mathcal{L}_{p}$ improves the mAP by 1.3\% to 21.8\% since it enforces reflectance interchangeability.

\noindent\underline{\textit{Decomposition Cohering Loss.}} We then utilize the proposed redecomposition cohering loss during training. This is denoted by $\mathcal{L}_{rc}$ under column DP in Table~\ref{table:ablation}. The proposed loss boosts the performance of not using decomposition process {from 20.5\% to 22.3\%}, which is superior to the image decomposition loss. This is because we fully exploit the reflectance interchangeability by redecomposing and cohering the sequentially decomposed results.

\noindent\underline{\textit{Simultaneous Usage.}} We can also use both losses. The result in Table~\ref{table:ablation} displays that it hardly brings improvement than solely using our redecomposition cohering loss. This indicates that the effect of decomposition loss is essentially covered by the redecomposition cohering loss. 

\noindent\underline{\textit{More Redecomposition. }}We have also tried to implement the decomposition three times. The performance is hardly improved. Considering the computation efficiency, we stick to the two-times decomposition.

\noindent\textbf{Mutual {Feature Alignment} Loss.} The mutual {feature alignment} loss {in Eq.~\ref{eq:mc}} exerts a constraint on illumination invariance on feature level. Under the column $\mathcal{L}_{mfa}$ in Table~\ref{table:ablation}, we show that $\mathcal{L}_{mfa}$ {using KL-Divergence as our original design} increases mAP {from 22.3\% to 23.5\%}, which validates its effectiveness.

\noindent\underline{\textit{Choice of loss functions.}} We further replace KL-Divergence  {in Eq.~\ref{eq:mc}} with L1- and L2-Distance, denoted by L1 and L2 in Table~\ref{table:ablation}. Both degrade performance because of their harsh constraints.

\noindent\textbf{Feature Visualization.} 
\underline{\textit{Feature map visualization.}} We extract {backbone features from {$g_f$}} in face detection model~\cite{li2019cvpr} \wrt a given image; then transform the features into a single-channel map by applying channel-wise averaging. {We visualize these maps in Fig.~\ref{fig:feat} to show the improvement of our method visually on DARK FACE; furthermore, we visualize them on real paired well-lit and low-light images from the LOL-v2 dataset~\cite{yang2020cvpr} for the observation of their reflectance consistency.} Compared with DSFD~\cite{li2019cvpr}, our feature map captures clearer and more detailed object information in extremely low-light environments. Though sharing the same architecture and workflow with DSFD during inference, reflectance-aware information is implicitly injected into ours, contributing to higher-quality features in the dark.

\noindent\underline{\textit{Feature channel visualization. }}
We calculate the 
mean magnitude of every feature channel from all $g_f$ features as \cite{luo2022iccv,cai2022nips}. Mean magnitudes in Fig.~\ref{fig:featstat} are calculated on well-lit samples (purple mesh) in WIDER FACE and low-light samples (green histograms) in DARK FACE.
In a vanilla DSFD trained on WIDER FACE, {the mean magnitudes of channels \wrt well-lit and low-light images are misaligned with each other.} Some channels of low-light features are suppressed compared with well-lit features. This illustrates how illumination degrades a detector from a feature channel perspective. In DAI-Net, the relative mean magnitudes are consistent across two sets, validating our network learns illumination-invariant information that does not change along illumination.

\noindent\textbf{Qualitative Examples.} Some qualitative results on DARK FACE and ExDark are given in Fig.~\ref{fig:result}. {More visualized results are given in the supplementary material.}

\section{Conclusion}

This paper studies dark object detection in a new setting named zero-shot dark 
domain generalization. We propose a novel DArk-Illuminated Network (DAI-Net) to learn illumination invariant representations from a well-lit source domain. We first introduce a reflectance representation learning module to extract illumination-invariant feature representations from well-lit images. In the module, a reflectance decoding branch and a {mutual feature alignment loss} are designed to enforce illumination invariance on both image- and feature-level. Next, we improve the Retinex image decomposition process by designing an interchange-redecomposition-coherence procedure. We perform two times sequential decomposition, where the input of the latter is obtained by interchanging reflectance outputs from the former. At last, a redecomposition cohering loss is designed to enforce the consistency between decomposition results over the two sequential decompositions. Experiments on DARK FACE and ExDark validate the strong dark domain generalizability of DAI-Net.

\newpage
{
    \small
    \bibliographystyle{ieeenat_fullname}
    \bibliography{main}
}

\clearpage
\setcounter{page}{1}
\maketitlesupplementary

This supplementary material provides 1) more implementation details; 2) additional ablation study; 
3) more examples for visualization and comparison.

\section{Implementation Details} We implement our method with PyTorch 1.11.0 and all experiments are conducted on four V100 GPUs, except for the image classification task we use a single V100 GPU.

\section{Ablation Study}

{The experiments below are performed on WIDER FACE $\rightarrow$ DARK FACE using the DSFD detector~\cite{li2019cvpr} as the same to Sec.~\ref{subsec:darkface} in the paper.}

\begin{table}[t]

\begin{minipage}{0.5\linewidth}
    \centering
    \vspace{0.48in}

	\begin{tabular}{c|c}
    
		\toprule
		($M,N$) & mAP(\%)\\
	\midrule
        ($M$=2, $N$=2) &\textbf{23.5}\\
        \midrule
        ($M$=1, $N$=2) & 21.9\\
        ($M$=4, $N$=2) & 22.3\\
        \midrule
        ($M$=2, $N$=1) & 22.4\\
        ($M$=2, $N$=4) & 20.3\\
        ($M$=2, $N$=7) & 18.6\\
    \bottomrule
	\end{tabular}

    	\caption{Parameter variation ($M, N$) in reflectance decoder.}
     
	\label{tab:ablationrefdec}

\end{minipage}
\hfill
\begin{minipage}{0.48\linewidth}
    \centering

    \begin{minipage}{\linewidth}
    \centering

      	\begin{tabular}{c|c}
    
		\toprule
		Method & mAP(\%)\\
	\midrule
        Ours & 23.5\\
        w/ RF & 24.2\\
    \bottomrule
	\end{tabular}
        
    % }
    \vspace{-0.03in}

    	\caption{Alternative of Retinex decomposition net.}
     \vspace{0.05in}
 
\label{tab:ablationreflosses}
    \end{minipage}
    
    \begin{minipage}{\linewidth}
    \centering

    	\begin{tabular}{c|c}   
		\toprule
		Method & mAP(\%)\\
	\midrule
        Baseline & 15.2\\
        % \midrule
        +$\mathcal{L}_{ref}$ & 18.9\\

        +$\mathcal{L}_{decom}$ & \textbf{20.5}\\
    \bottomrule
	\end{tabular}
 
    % }

    	\caption{Analysis on reflectance decoding losses.}

\label{tab:ablationdecomnet}  
    \end{minipage}
\end{minipage}
\vspace{-0.17in}
\end{table}

\noindent\textbf{More results for reflectance decoder.} In the paper, we report that the best performance occurs when the reflectance decoder is appended after the $M$-th conv layer of backbone and it consists of $N$ conv layers. Both $M$ and $N$ are set to 2 by default. Here, we conduct parameter variation over different $M$ and $N$ in Table~\ref{tab:ablationrefdec}.

\noindent\underline{\textit{Variation of M.}} We first fix $N$ and conduct experiments on different $M$. The result shows that either increasing or decreasing $M$ would slightly degrade the performance. $M=2$ performs the best.

\begin{figure}[t]

% \vspace{-10px}
% \begin{overpic} 
% [width=\linewidth]
% {example-image-a}
% \end{overpic}
\centering
\includegraphics[width=0.95\linewidth]{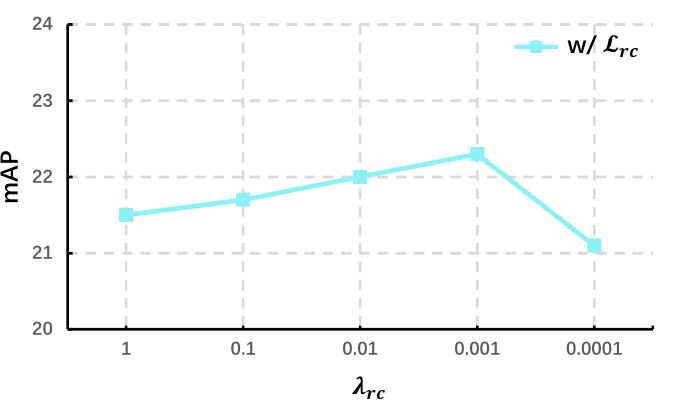}

\caption{ Parameter variation of loss weight $\lambda_{rc}$ for redecomposition cohering loss.
}
% \vspace{-0.125in}
\label{fig:lossweight}
\vspace{-0.15in}
\end{figure}

\noindent\underline{\textit{Variation of N.}} Next, we fix $M$ to 2 and vary the value of $N$ to find whether a bigger $N$ benefits the reflectance decoder. When $N$ increases, the performance actually drops. The default one $N=2$ performs the best.

\noindent\underline{\textit{Alternative of the Retinex decomposition net.}} {The Retinex decomposition net (see Sec.~\ref{sec:train} and Fig.~\ref{fig:overview} in the paper) can be replaced with other methods with similar functions. Here, we replace it with a recent development called RetinexFormer~\cite{cai2023iccv} in the paper. RetinexFormer explicitly models noises in reflectance and illumination maps through a transformer architecture, thus offering more robust pseudo ground truth than the original RetinexNet~\cite{wei2018bmvc} used in the paper. We achieve better result by using RetinexFormer (w/ RF) in Table~\ref{tab:ablationreflosses}. This illustrates the generalizability of our method with a stronger decomposition net.}

\noindent\underline{\textit{Analysis of the reflectance decoding losses.}} {Referring to Sec.~\ref{sec:train}-DAI-Net in the paper, for the optimization of reflectance decoder, we also add the reflectance learning loss $\mathcal{L}_{ref}$ and image decomposition loss $\mathcal{L}_{decom}$. 
In Table~\ref{tab:ablationdecomnet}, we give an analysis by adding these losses sequentially to our Baseline (see Sec.~\ref{sec:ab} in the paper). Every employed loss contributes clearly to the overall performance.}

\noindent\textbf{More results for redecomposition cohering loss.} We vary the loss weight $\lambda_{rc}$ for redecomposition cohering loss in Fig.~\ref{fig:lossweight}. Ranging from 0.0001 to 1 for $\lambda_{rc}$, we observe that using the redecomposition cohering loss consistently shows better performance compared to not using it (20.5 in Table~\ref{table:ablation} in the paper). $\lambda_{rc}=0.001$ achieves the largest improvement.

\section{Visualization}

\begin{figure*}[t]

% \begin{overpic} 
% [width=\linewidth]
% {example-image-a}
% \end{overpic}
\centering
\includegraphics[width=0.8\textwidth]{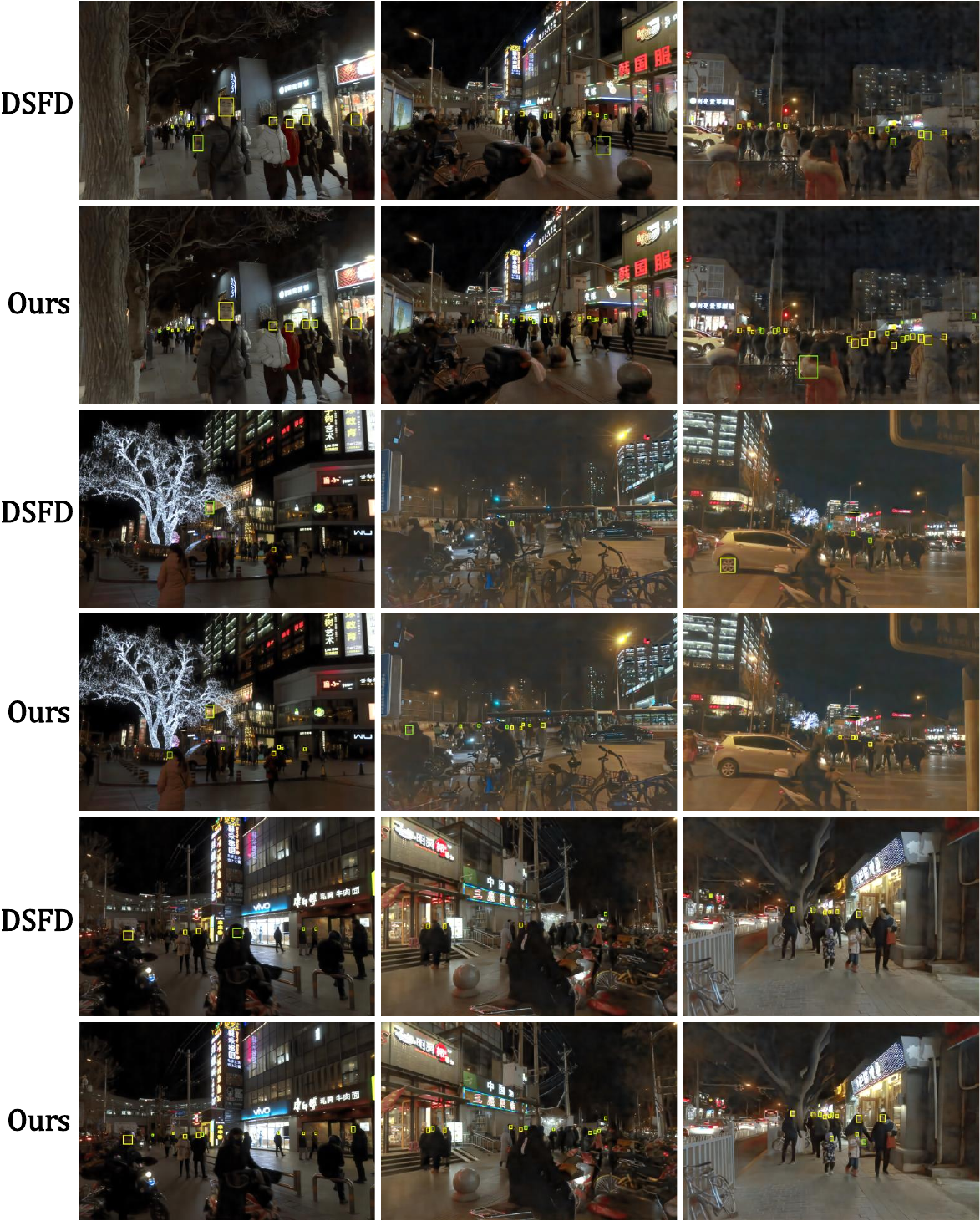}

\caption{ Qualitative examples of DSFD~\cite{li2019cvpr} and our method. Images are taken from the DARK FACE dataset and are enhanced only for visualization.
}
\label{fig:suppdf}

\end{figure*}

\begin{figure*}[t]

% \begin{overpic} 
% [width=\linewidth]
% {example-image-a}
% \end{overpic}
\centering
\includegraphics[width=0.75\textwidth]{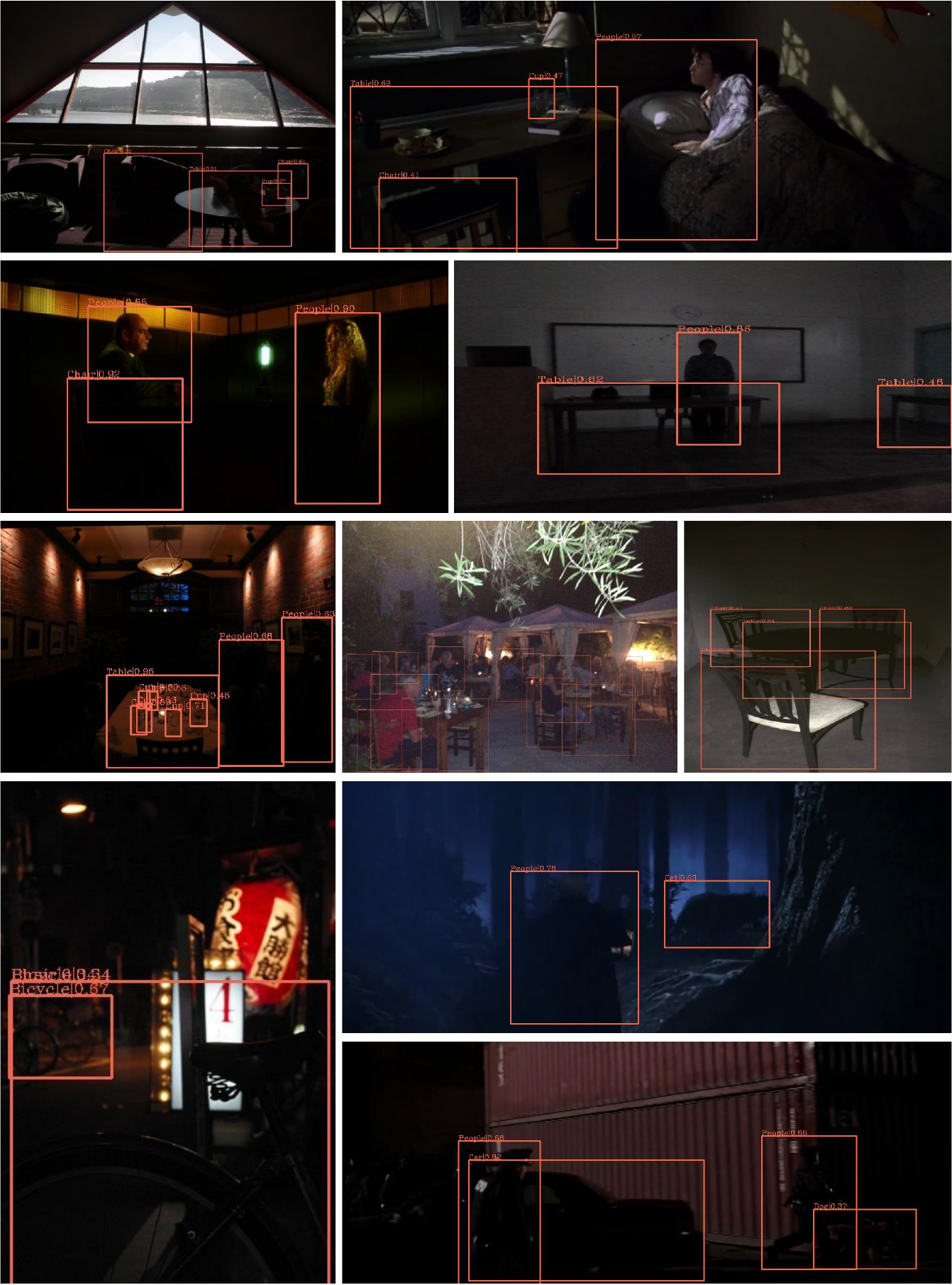}

\caption{ Qualitative examples of our method on the ExDark dataset. Predicted categories and their confidence scores are given along with bounding boxes in each image. Zoom in for details. 
}
\label{fig:suppex}

\end{figure*}

\noindent\textbf{Qualitative examples.}
We present additional qualitative examples of our method on {WIDER FACE $\rightarrow$ DARK FACE and COCO $\rightarrow$ ExDark} in Fig.~\ref{fig:suppdf} and Fig.~\ref{fig:suppex}, respectively. 

\begin{figure*}[!t]

\centering
\includegraphics[width=0.75\linewidth]{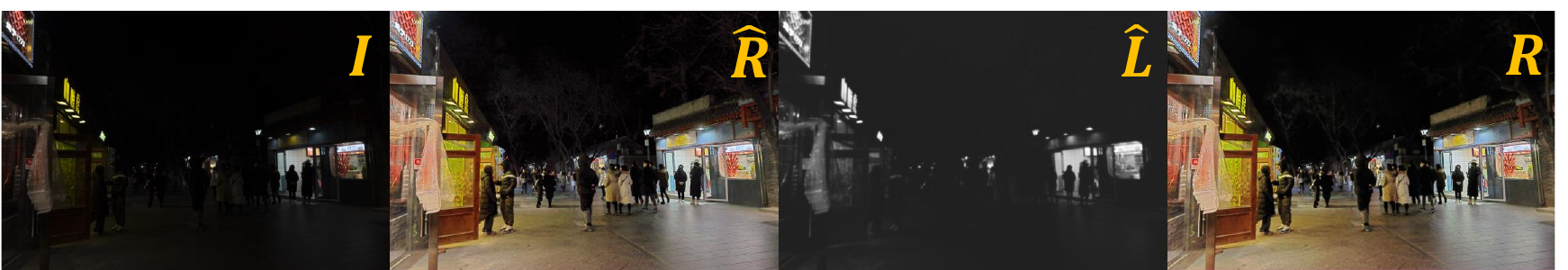}
% \vspace{-0.2in}
    	\vspace{-5px}

\caption{ 
Visualization of original image, pseudo ground truth, and predicted reflectance (from left to right).
}
\vspace{-18px}
\label{fig:decomp}

\end{figure*}

\begin{figure*}[hb]

% \begin{overpic} 
% [width=\linewidth]
% {example-image-a}
% \end{overpic}
\centering
\includegraphics[width=0.7\textwidth]{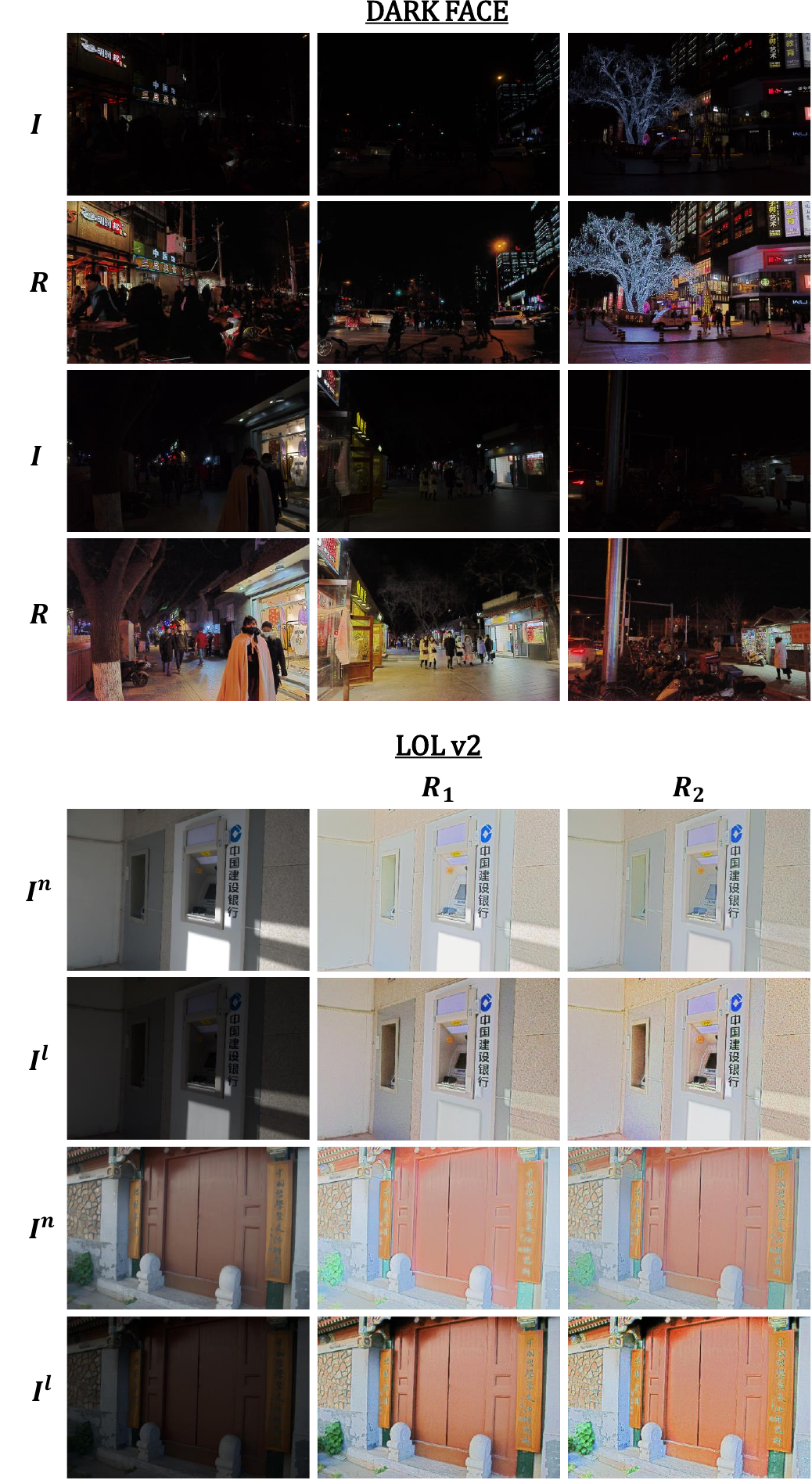}

\caption{ Reflectance visualization. We show the predicted reflectances $R$ over extreme low-light images $I$ from the DARK FACE dataset. We also give two-round decomposed reflectances $R_1^n,R_1^l,R_2^n,R_2^l$ of real paired images $I^n,I^l$ from LOL v2.
}
\label{fig:suppref}

\end{figure*}

\noindent\textbf{Reflectance visualization.} {Given an image $I$ from the DARK FACE dataset, we show the {pseudo ground truth $\{\hat{R},\hat{L}\}$ and the }decomposed reflectance $R$ from the reflectance decoder of our DAI-Net {in Fig.~\ref{fig:decomp}. More visualizations of $R$ are given} in Fig.~\ref{fig:suppref} (Top). We also provide the two-round decomposition results on LOL v2. We show them in Fig.~\ref{fig:suppref} (Bottom). The generated reflectances are consistent between well-lit and low-light image pairs, and also between the two rounds.}

\end{document}